\documentclass[11pt,a4paper]{article}
\usepackage[hyperref]{acl2019}
\usepackage{times}
\usepackage{latexsym}
\usepackage{url}
\usepackage{CJKutf8}

\usepackage{epsfig,graphicx,caption,subcaption}
\usepackage{amsmath,amssymb}
\usepackage{multirow,booktabs, hhline}

\aclfinalcopy 

\title{OpenHowNet: An Open Sememe-based Lexical Knowledge Base}

\author{Fanchao Qi$^1$, Chenghao Yang$^2$\thanks{\ \ Work done while doing internship at Tsinghua University.}\hspace{0.5em}, Zhiyuan Liu$^1$, Qiang Dong$^3$, Maosong Sun$^1$, Zhendong Dong$^3$\\
$^1$Department of Computer Science and Technology, Tsinghua University\\
  Institute for Artificial Intelligence, Tsinghua University\\
  State Key Lab on Intelligent Technology and Systems, Tsinghua University\\
$^{2}$Software College, Beihang University\\
$^{3}$HowNet Technology Inc.\\
}

\date{}

\begin{document}
\begin{CJK}{UTF8}{gkai}

\maketitle
\begin{abstract}
  In this paper, we present an open sememe-based lexical knowledge base OpenHowNet. Based on well-known HowNet, OpenHowNet comprises three components: core data which is composed of more than 100 thousand senses annotated with sememes, OpenHowNet Web which gives a brief introduction to OpenHowNet as well as provides online exhibition of OpenHowNet information, and OpenHowNet API which includes several useful APIs such as accessing OpenHowNet core data and drawing sememe tree structures of senses. In the main text, we first give some backgrounds including definition of sememe and details of HowNet. And then we introduce some previous HowNet and sememe-based research works. Last but not least, we detail the constituents of OpenHowNet and their basic features and functionalities. Additionally, we briefly make a summary and list some future works.
\end{abstract}

\section{Introduction}

In the field of Natural Language Processing (NLP), words are generally the smallest objects of study because they are considered as the smallest meaningful units that can stand by themselves of human languages. However, the meanings of words can be further divided into smaller parts. For example, the meaning of ``man" can be considered as the combination of the meanings of ``human'',``male'' and ``adult'', and the meaning of ``boy'' is composed of the meanings of ``human'', ``male'' and ``child''. In linguistics, the minimum indivisible units of meaning, i.e. semantic units, are defined as sememes \citep{bloomfield1926set}. And some linguists believe that meanings of all the words can be composed of a limited closed set of sememes.

However, sememes are implicit and as a result, it is hard to intuitively define the set of sememes and determine which sememes a word have at a glance. Therefore, some researchers spend tens of years sifting sememes from all kinds of dictionaries and linguistic knowledge bases (KBs), and annotating words with these selected sememes to construct sememe-based linguistic KB. HowNet \citep{dong2003hownet} is the most famous one of such KBs.

\section{HowNet}

\begin{figure}[!h]
   \centering
   \begin{subfigure}{0.52\linewidth}
      \centering
      \includegraphics[width=\linewidth]{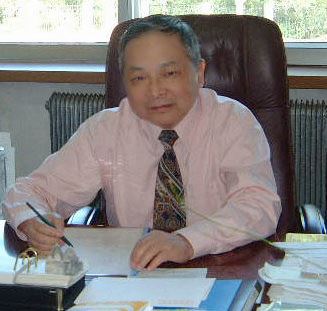}
      \caption{Zhendong Dong}
   \end{subfigure}%
   \begin{subfigure}{0.47\linewidth}
      \centering
      \includegraphics[width=\linewidth]{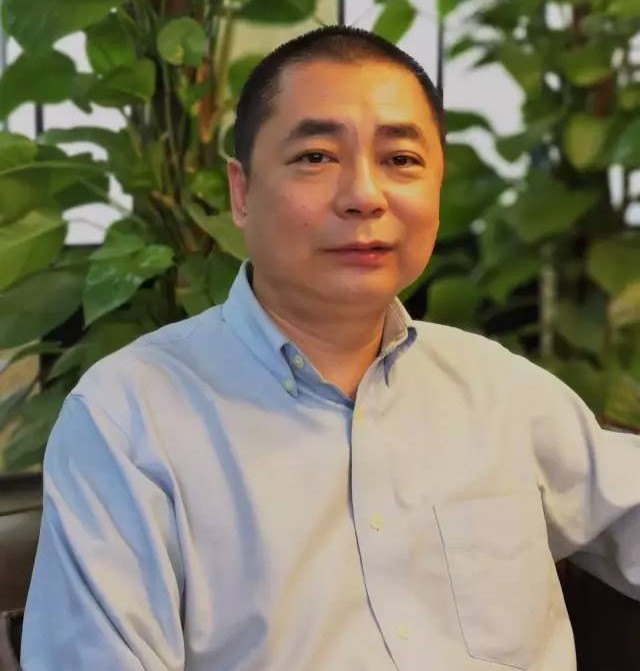}
      \caption{Qiang Dong}
   \end{subfigure}%
\caption{Founders of HowNet}
\label{fig:founder}
\end{figure}

HowNet was initially designed and constructed by Zhendong Dong and his son Qiang Dong (Figure \ref{fig:founder}) in the 1990s. And it has kept frequent updating since it was published in 1999. 

The sememe set of HowNet is determined by extracting, analyzing, merging and filtering semantics of thousands of Chinese characters. And the sememe set can also be adjusted or expanded in the subsequent process of annotating words. Each sememe in HowNet is represented by a word or phrase in Chineses and English such as \texttt{human|人} and \texttt{ProperName|专}.

\begin{figure*}[!t]
	\includegraphics[width=\linewidth]{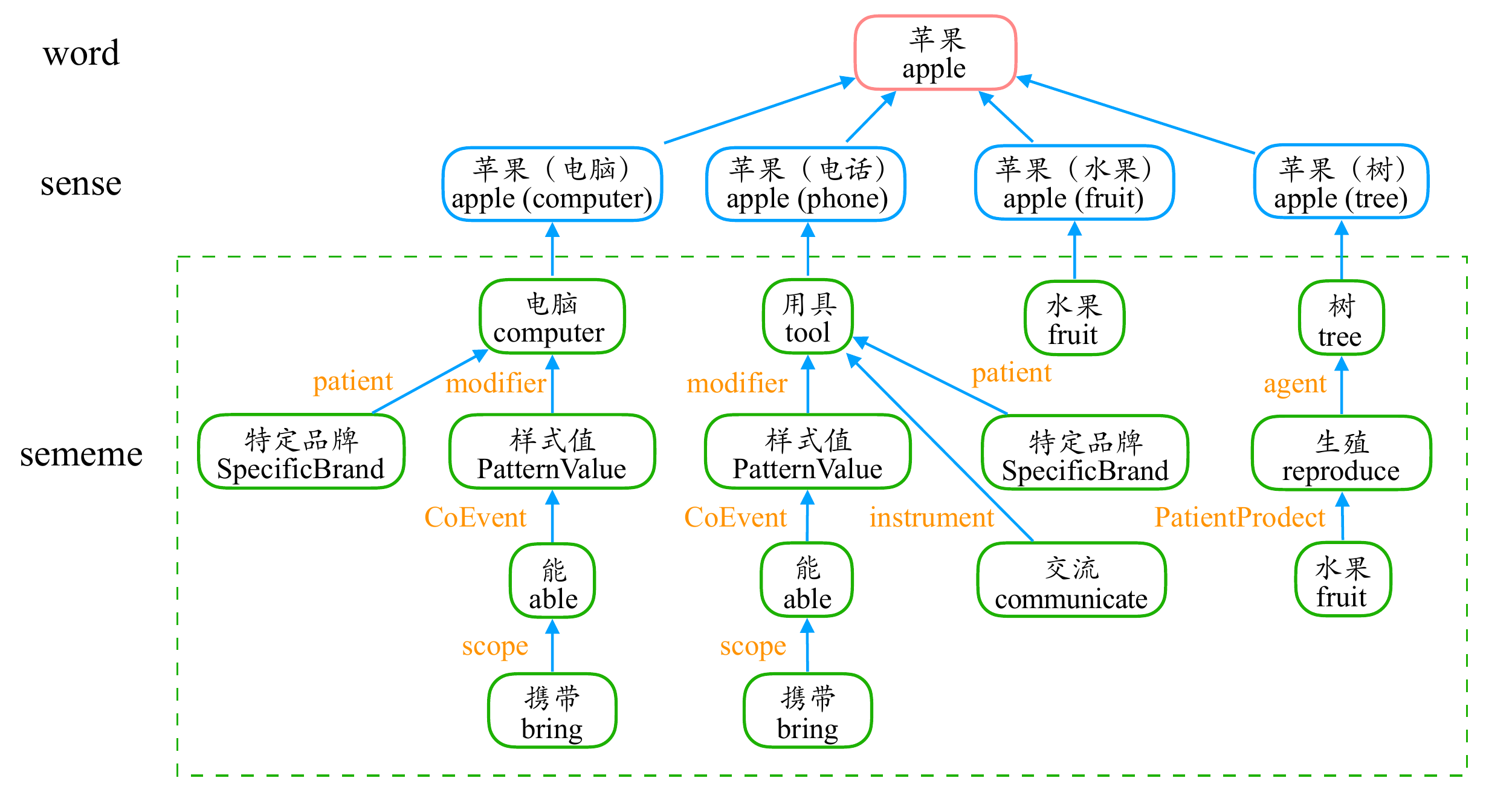}
\caption{An example of word annotated with sememes in HowNet}
\label{fig:example}
\end{figure*}

HowNet also builds a taxonomy for the sememes. All the sememes of HowNet can be classified as one of the following types: Thing, Part, Attribute, Time, Space, Attribute Value and Event. In addition, to depict the semantics of words more precisely, HowNet incorporates relations between sememes, which are called ``dynamic roles'', into the sememe annotations of words. 

Considering the polysemy, HowNet differentiates diverse senses of each word in the sememe annotations. And each sense is also expressed in both Chinese and English. An example of sememe annotation for a word is illustrated in Figure \ref{fig:example}. We can see from the figure that the word ``apple'' has four senses including ``apple (computer)'', ``apple (phone)'', ``apple (fruit)'' and ``apple (tree)'', and each sense is the root node of a ``sememe tree'' where any pair of father and son sememe node is multi-relational. 

Additionally, HowNet annotates POS tag for each sense, and add sentiment category as well as some usage examples for certain senses.  

The latest version of HowNet is published in January 2019 and the statistics are shown in Table \ref{table:statistics}.

\begin{table}[!h]
\small
\centering{
\begin{tabular}{cccc}
  \toprule
  Type & Count \\
  \midrule
  Sense & 229,767\\
  Distinct Chinese word & 127,266 \\
  Distinct English word & 104,025 \\
  Sememe & 2,187 \\
  \bottomrule
\end{tabular}
}
\caption{Statistics of latest version of HowNet}
\label{table:statistics}
\end{table}

\section{HowNet and Sememe-related Researches}

Since HowNet was published, it has attracted wide attention. People use HowNet and sememe in various NLP tasks including word similarity computation \citep{liu2002}, word sense disambiguation \citep{zhang2005chinese}, question classification \citep{sun2007hownet} and sentiment analysis \citep{dang2010method,xianghua2013multi}. Among these researches, \citet{liu2002} is one of the most influential works, in which similarities of given two words are computed by measuring the degree of resemblance of their sememe trees.

Recent years also witness some works incorporating sememes into neural network models. \citet{niu2017improved} propose a novel word representation learning model named SST that reforms Skip-gram \citep{mikolov2013efficient} by adding contextual attention to senses of target word, which are represented with combinations of corresponding sememes' embeddings. Experimental results show that SST can not only improve the quality of word embeddings but also learn satisfactory sense embeddings to do word sense disambiguation.

\citet{gu2018language} incorporate sememes into the decoding phase of language modeling where sememes are predicted first, and then senses and words are predicted in succession. The proposed model shows enhancement in the perplexity of language modeling and performance of downstream task headline generation.

Besides, HowNet is also utilized in lexicon expansion \citep{zeng2018chinese}, semantic rationality evaluation \citep{Liu2018}, etc.

Considering that human annotation is time-consuming and labor-intensive, some works attempt to employ machine learning methods to predict sememes for new words automatically. \citet{xie2017lexical} propose the task firstly and present two simple but effective models SPWE, which is based on collaborative filtering and SPSE, which is based on matrix factorization. \citet{jin2018incorporating} further take the internal information of words into account when predicting sememes and achieve considerable boost of performance. And \citet{Li2018Sememe} take advantage of definitions of words to predict sememes. As for \citet{Qi2018Cross}, they propose the task of cross-lingual lexical sememe prediction and present a bilingual word representation learning and alignment-based model which demonstrates effectiveness in predicting sememes for cross-lingual words. 

\section{OpenHowNet}

With the support of founders of HowNet, OpenHowNet encompasses the core data of HowNet and provides free download. In addition, OpenHowNet comprises OpenHowNet Web and OpenHowNet API.

\begin{figure}[!ht]
	\includegraphics[width=\linewidth]{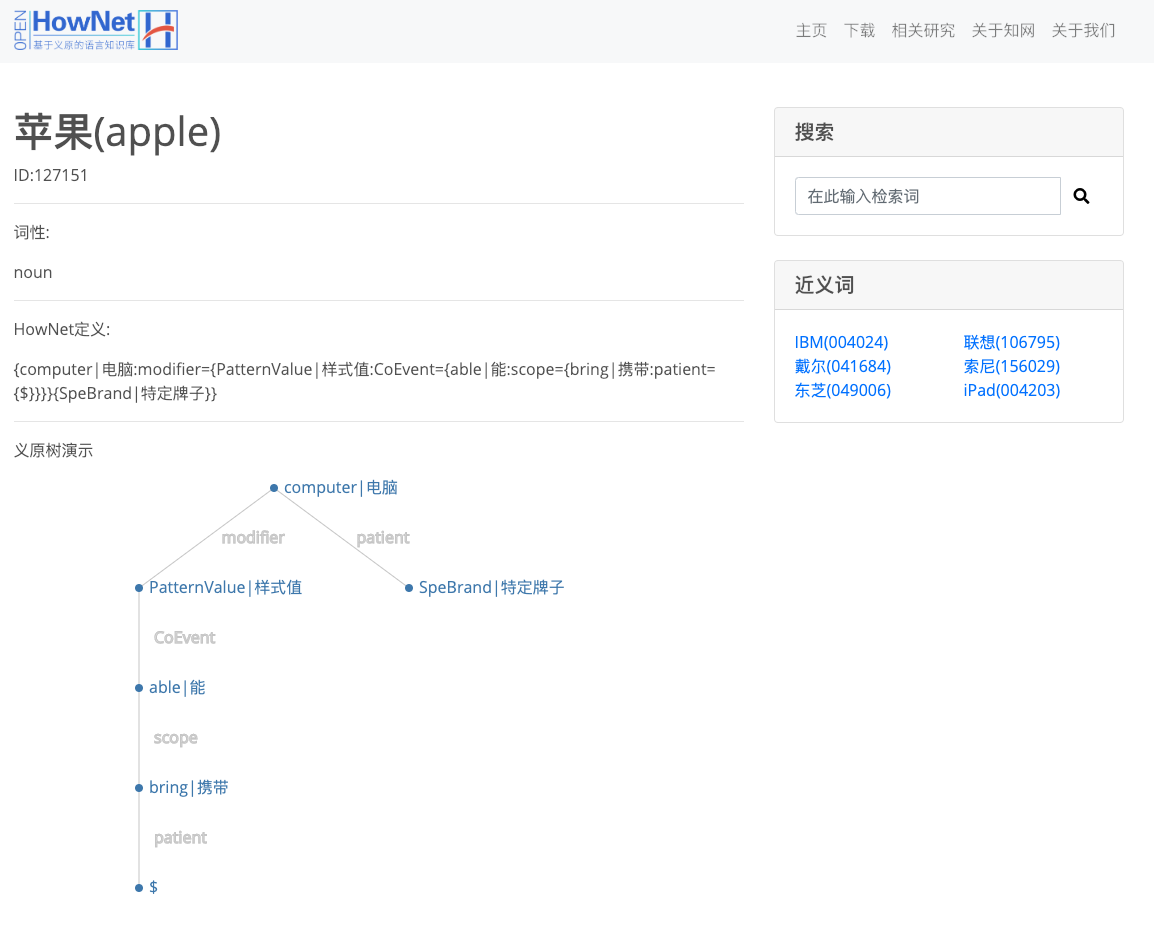}
\caption{An example of sense information illustration on OpenHowNet Web}
\label{fig:open}
\end{figure}

OpenHowNet Web\footnote{\url{https://openhownet.thunlp.org/home}} gives a concise introduction to OpenHowNet, provides statistics of OpenHowNet dataset, lists sememe-related researches as well as history and founders of HowNet. Moreover, it has the functions of sense retrieval and sememe tree demonstration. Figure \ref{fig:open} shows how OpenHowNet Web illustrates the information of a sense in HowNet. 
From the figure, we can see that the exhibited information of a sense includes sense ID, POS tag, sememe-based definition, sememe tree and semantically near senses, which are computed with the sememe-based word similarity algorithm proposed by \citet{liu2002}.

Also, on the website of OpenHowNet, you can apply to download the whole OpenHowNet dataset and semantic representation learning results, which are learned by SST model \citep{niu2017improved} and include word, sense and sememe embeddings. Downloading is totally free and you just need to fill in a simple form including email address in order to inform you of update of OpenHowNet.

OpenHowNet API\footnote{\url{https://github.com/thunlp/OpenHowNet-API}} provides some useful APIs including searching senses and sememes from OpenHowNet dataset, drawing sememe tree of a given sense, computing word similarity based on sememe tree comparison \citep{liu2002}, etc. There are documents describing how to uses these APIs and you can also try the Jupyter Notebook tutorial. 

\section{Conclusion and Future Work}

In this paper we present OpenHowNet, an open sememe-based lexical knowledge base which is based on well-known HowNet. We give an introduction to sememes, philosophy and structure of HowNet, sememe-related researches as well as basic components of OpenHowNet and their functions.

In the future, we will try to utilize machine learning methods to check and improve the annotation consistency of OpenHowNet. Moreover, we will attempt to ensemble various automatic sememe prediction methods together with online interactive annotating to annotate more words with sememes and enlarge the scale of OpenHowNet. Also, we want to transfer the existing sememe knowledge to other languages and build a multi-lingual version of OpenHowNet.

\bibliography{reference}

\begin{thebibliography}{16}
\expandafter\ifx\csname natexlab\endcsname\relax\def\natexlab#1{#1}\fi

\bibitem[{Bloomfield(1926)}]{bloomfield1926set}
Leonard Bloomfield. 1926.
\newblock A set of postulates for the science of language.
\newblock \emph{Language}, 2(3):153--164.

\bibitem[{Dang and Zhang(2010)}]{dang2010method}
Lei Dang and Lei Zhang. 2010.
\newblock Method of discriminant for chinese sentence sentiment orientation
  based on hownet.
\newblock \emph{Application Research of Computers}, 4:43.

\bibitem[{Dong and Dong(2003)}]{dong2003hownet}
Zhendong Dong and Qiang Dong. 2003.
\newblock Hownet: a hybrid language and knowledge resource.
\newblock In \emph{Proceedings of NLP-KE}.

\bibitem[{Fu et~al.(2013)Fu, Liu, Guo, and Wang}]{xianghua2013multi}
Xianghua Fu, Guo Liu, Yanyan Guo, and Zhiqiang Wang. 2013.
\newblock Multi-aspect sentiment analysis for chinese online social reviews
  based on topic modeling and hownet lexicon.
\newblock \emph{Knowledge-Based Systems}, 37:186--195.

\bibitem[{Gu et~al.(2018)Gu, Yan, Zhu, Liu, Xie, Sun, Lin, and
  Lin}]{gu2018language}
Yihong Gu, Jun Yan, Hao Zhu, Zhiyuan Liu, Ruobing Xie, Maosong Sun, Fen Lin,
  and Leyu Lin. 2018.
\newblock Language modeling with sparse product of sememe experts.
\newblock In \emph{Proceedings of EMNLP}.

\bibitem[{Jin et~al.(2018)Jin, Zhu, Liu, Xie, Sun, Lin, and
  Lin}]{jin2018incorporating}
Huiming Jin, Hao Zhu, Zhiyuan Liu, Ruobing Xie, Maosong Sun, Fen Lin, and Leyu
  Lin. 2018.
\newblock Incorporating chinese characters of words for lexical sememe
  prediction.
\newblock In \emph{Proceedings of ACL}.

\bibitem[{Li et~al.(2018)Li, Ren, Dai, Wu, Wang, and Sun}]{Li2018Sememe}
Wei Li, Xuancheng Ren, Damai Dai, Yunfang Wu, Houfeng Wang, and Xu~Sun. 2018.
\newblock {Sememe Prediction: Learning Semantic Knowledge from Unstructured
  Textual Wiki Descriptions}.
\newblock \emph{arXiv preprint}.

\bibitem[{Liu and Li(2002)}]{liu2002}
Qun Liu and Sujian Li. 2002.
\newblock Word similarity computing based on hownet.
\newblock \emph{International Journal of Computational Linguistics \& Chinese
  Language Processing}, 7(2):59--76.

\bibitem[{Liu et~al.(2018)Liu, Xu, Ren, and Sun}]{Liu2018}
Shu Liu, Jingjing Xu, Xuancheng Ren, and Xu~Sun. 2018.
\newblock Evaluating semantic rationality of a sentence: A sememe-word-matching
  neural network based on hownet.
\newblock \emph{arXiv preprint}.

\bibitem[{Mikolov et~al.(2013)Mikolov, Chen, Corrado, and
  Dean}]{mikolov2013efficient}
Tomas Mikolov, Kai Chen, Greg Corrado, and Jeffrey Dean. 2013.
\newblock Efficient estimation of word representations in vector space.
\newblock In \emph{Proceedings of ICLR}.

\bibitem[{Niu et~al.(2017)Niu, Xie, Liu, and Sun}]{niu2017improved}
Yilin Niu, Ruobing Xie, Zhiyuan Liu, and Maosong Sun. 2017.
\newblock Improved word representation learning with sememes.
\newblock In \emph{Proceedings of ACL}.

\bibitem[{Qi et~al.(2018)Qi, Lin, Sun, Zhu, Xie, and Liu}]{Qi2018Cross}
Fanchao Qi, Yankai Lin, Maosong Sun, Hao Zhu, Ruobing Xie, and Zhiyuan Liu.
  2018.
\newblock {Cross-lingual Lexical Sememe Prediction}.
\newblock In \emph{Proceedings of EMNLP}.

\bibitem[{Sun et~al.(2007)Sun, Cai, Lv, and Dong}]{sun2007hownet}
Jingguang Sun, Dongfeng Cai, Dexin Lv, and Yanju Dong. 2007.
\newblock Hownet based chinese question automatic classification.
\newblock \emph{Journal of Chinese Information Processing}, 21(1):90--95.

\bibitem[{Xie et~al.(2017)Xie, Yuan, Liu, and Sun}]{xie2017lexical}
Ruobing Xie, Xingchi Yuan, Zhiyuan Liu, and Maosong Sun. 2017.
\newblock Lexical sememe prediction via word embeddings and matrix
  factorization.
\newblock In \emph{Proceedings of AAAI}.

\bibitem[{Zeng et~al.(2018)Zeng, Yang, Tu, Liu, and Sun}]{zeng2018chinese}
Xiangkai Zeng, Cheng Yang, Cunchao Tu, Zhiyuan Liu, and Maosong Sun. 2018.
\newblock Chinese liwc lexicon expansion via hierarchical classification of
  word embeddings with sememe attention.
\newblock In \emph{Proceedings of AAAI}.

\bibitem[{Zhang et~al.(2005)Zhang, Gong, and Wang}]{zhang2005chinese}
Yuntao Zhang, Ling Gong, and Yongcheng Wang. 2005.
\newblock Chinese word sense disambiguation using hownet.
\newblock In \emph{Proceedings of International Conference on Natural
  Computation}.

\end{thebibliography}
\bibliographystyle{acl_natbib}

\end{CJK}
\end{document}